\documentclass{article}

\usepackage{arxiv}

\usepackage[utf8]{inputenc} 
\usepackage[T1]{fontenc}    
\usepackage{hyperref}       
\usepackage{url}            
\usepackage{booktabs}       
\usepackage{amsfonts}       
\usepackage{nicefrac}       
\usepackage{microtype}      
\usepackage[sorting=none]{biblatex}
\usepackage[frozencache,cachedir=.]{minted}
\usepackage{graphicx}
\usepackage{xcolor}
\definecolor{LightGray}{gray}{0.95}
\definecolor{codegreen}{rgb}{0,0.6,0}
\definecolor{codegray}{rgb}{0.5,0.5,0.5}
\definecolor{codepurple}{rgb}{0.58,0,0.82}
\definecolor{backcolour}{rgb}{0.95,0.95,0.92}

\definecolor{codegray2}{gray}{0.9}
\newcommand{\code}[1]{\colorbox{codegray2}{\texttt{#1}}}

\newcommand\gcam{0.2}
\newcommand\x{0.25}

\title{M3d-CAM: A Pytorch library to generate 3D attention maps for medical deep learning}

\author{
  Karol Gotkowski \\
  Graphisch-Interaktive Systeme\\
  Technical university of Darmstadt\\
  \And
  Camila Gonzalez \\
  Graphisch-Interaktive Systeme\\
  Technical university of Darmstadt\\
      \And
  \hspace{-1.3cm}Andreas Bucher \\
  \hspace{-1.3cm}Institut für Diagnostische und Interventionelle Radiologie\\
  \hspace{-1.3cm}Universitätsklinikum Frankfurt\\
    \And 
  \hspace{-1.4cm}Anirban Mukhopadhyay \\
  \hspace{-1.4cm}Graphisch-Interaktive Systeme\\
  \hspace{-1.0cm}Technical university of Darmstadt\\
}

\begin{document}
\maketitle

\begin{abstract}

M3d-CAM is an easy to use library for generating attention maps of CNN-based Pytorch models improving the interpretability of model predictions for humans. The attention maps can be generated with multiple methods like Guided Backpropagation, Grad-CAM, Guided Grad-CAM and Grad-CAM++. These attention maps visualize the regions in the input data that influenced the model prediction the most at a certain layer. Furthermore, M3d-CAM supports 2D and 3D data for the task of classification as well as for segmentation. A key feature is also that in most cases only a single line of code is required for generating attention maps for a model making M3d-CAM basically plug and play.
\end{abstract}


\section{M3d-CAM Overview}

\begin{figure}[!ht]
  \centering
  \begin{minipage}[b]{\gcam\textwidth}
    \includegraphics[width=\textwidth]{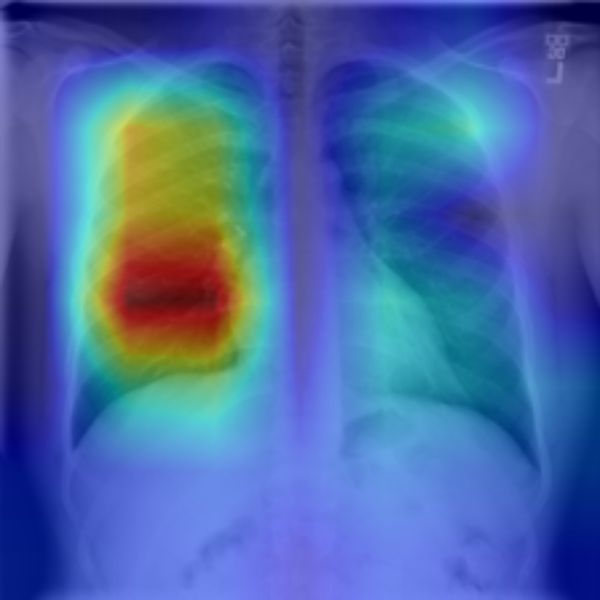}
  \end{minipage}
  \hspace{0.5cm}
  \begin{minipage}[b]{\gcam\textwidth}
    \includegraphics[width=\textwidth]{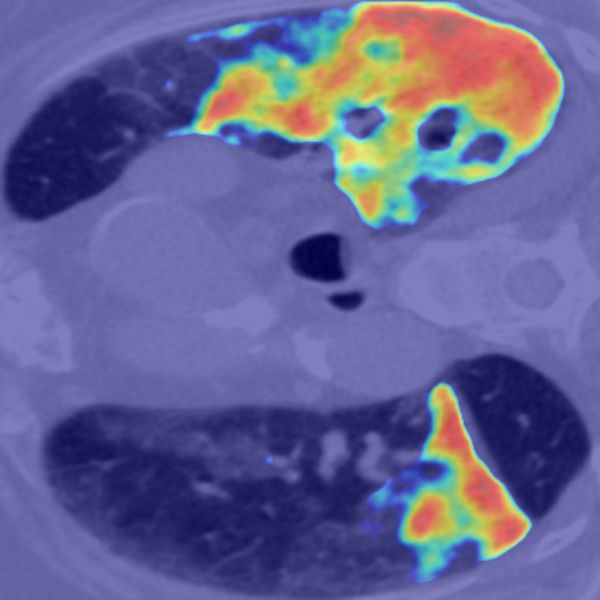}
  \end{minipage}
    \hspace{0.5cm}
  \begin{minipage}[b]{\gcam\textwidth}
    \includegraphics[width=\textwidth]{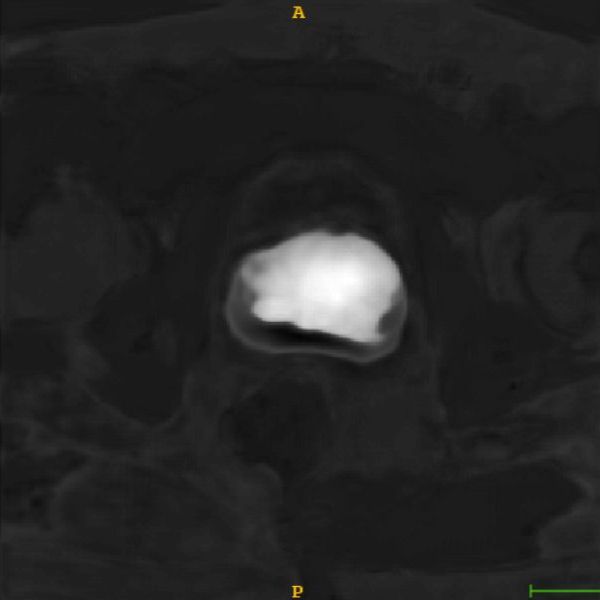}
  \end{minipage}
    \caption{Grad-CAM attention maps for 2D classification, 2D segmentation and 3D segmentation.}
  \label{fig:gcam0}
\end{figure}

M3d-CAM is an easy to use library for generating attention maps with any CNN-based Pytorch \cite{pytorch} model both for 2D and 3D data as well as with classification and segmentation tasks. M3d-CAM works by injecting itself into a given model appending and even replacing certain functions of the model. The model itself will work as usual and its predictions remain untouched ensuring that no code is broken. M3d-CAM itself will work behind the scenes and generate attention maps every time \code{model.forward} is called. Examples of these attention maps are shown in figure \ref{fig:gcam0}. The most important functions of M3d-CAM are explained in the following subsections. 

\subsection{Injection}
To inject a model with M3d-CAM one simply needs to insert the line \code{model = medcam.inject(model)} after model initialization as shown in code example \ref{code}. This will add all the necessary functionality to the model. Additionally \code{inject} offers multiple parameters that can be adjusted. As an example one can define an \code{output\_dir} and set \code{save\_maps=True} to save every generated attention map. One can also set a desired \code{backend} which is used for generating the attention maps such as Grad-CAM. These backends are explained in more detail in section \ref{backends}. Furthermore, it is possible to choose the layer of interest with \code{layer}. Hereby one can specifically define a single layer, a set of layers, every layer with  \textit{full} or the highest CNN-layer with \textit{auto} for the most comfort.

\begin{listing}[ht]
\begin{minted}
[
frame=lines,
framesep=2mm,
baselinestretch=1.2,
bgcolor=LightGray,
fontsize=\footnotesize,
linenos
]
{python}

# Import M3d-CAM
from medcam import medcam

# Init your model and dataloader
model = MyCNN()
data_loader = DataLoader(dataset, batch_size=1, shuffle=False)

# Inject model with M3d-CAM
model = medcam.inject(model, output_dir="attention_maps", save_maps=True)

# Continue to do what you're doing...
# In this case inference on some new data
model.eval()
for i, batch in enumerate(data_loader):
    # Every time forward is called, attention maps will be generated and saved
    output = model(batch)
    # more of your code...
\end{minted}
\caption{Example of injecting a model with M3d-CAM}
\label{code}
\end{listing}

\subsection{Layer retrieval}
As the layer names of a model are often unknown to the user, M3d-CAM offers the method \code{medcam.get\_layers(model)} for quickly acquiring every layer name of a model. However it needs to be noted that attention maps can not be generated for every type of layer. This is true for layer types such as fully connected, bounding box or other special types of layers. The attention for theses layers can be computed but it is not possible to project them back to the original input data, hence no attention maps can be generated.

\subsection{Evaluation}
M3d-CAM also supports the evaluation of attention maps with given ground truth masks by simply calling \code{model.forward(input, mask)} including the mask in the forward call. The attention map is then internally evaluated by the \code{medcam.Evaluator} class with a predefined metric by the user. Alternatively one can call the \code{medcam.Evaluator} class directly. By calling \code{model.dump()} or respectively \code{medcam.Evaluator.dump()} the evaluation results are saved as an excel table.

\newpage

\section{Backends} \label{backends}

\begin{figure}[!ht]
  \centering
  \begin{minipage}[b]{\x\textwidth}
    \includegraphics[width=\textwidth]{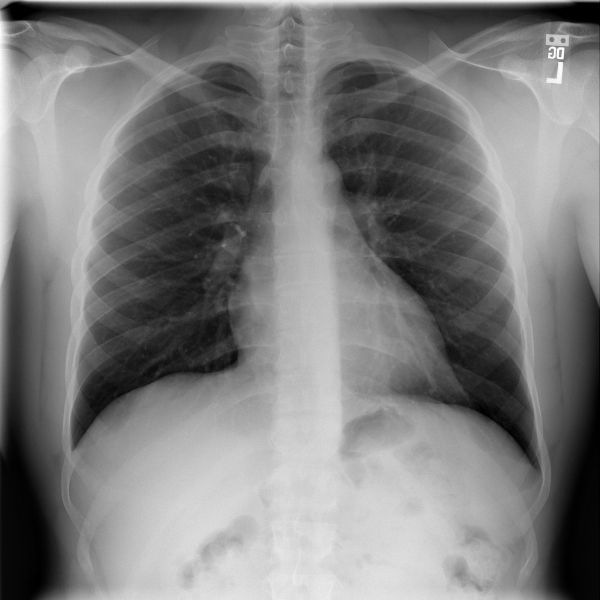}
  \end{minipage}
  \hspace{0.5cm}
  \begin{minipage}[b]{\x\textwidth}
    \includegraphics[width=\textwidth]{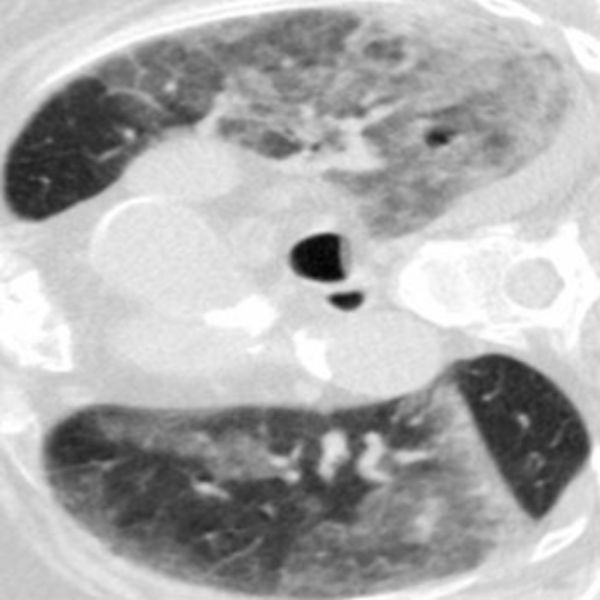}
  \end{minipage}
    \hspace{0.5cm}
  \begin{minipage}[b]{\x\textwidth}
    \includegraphics[width=\textwidth]{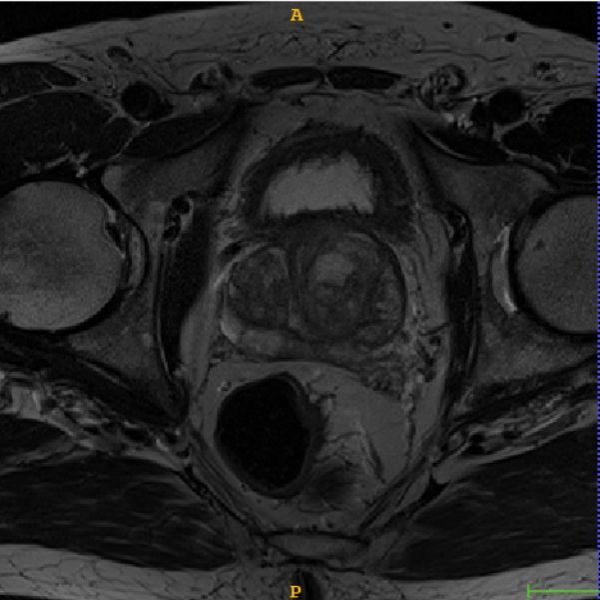}
  \end{minipage}
  \caption{From left to right: A chest X-Ray from the COVID-19 image data collection \cite{covid19}, a lung CT slice also from \cite{covid19} and 3D prostate CT image from the Medical Decathlon dataset \cite{prostate}}
  \label{fig:original}
\end{figure}

M3d-CAM supports multiple methods for generating the attention maps. For simplicity we will refer to them as \textit{backends}. For a better understanding of how these attention maps look like we included examples for every backend. The original input images are shown in figure \ref{fig:original}. The first image displays a chest X-Ray used on the task of classification by employing a CovidNet \cite{covidnet}, the second a lung CT slice on the task of 2D segmentation by employing an Inf-Net \cite{infnet} and the third a 3D prostate CT image on the task of 3D segmentation by employing a nnUNet \cite{nnunet}.

\subsection{Grad-CAM}
\begin{figure}[!ht]
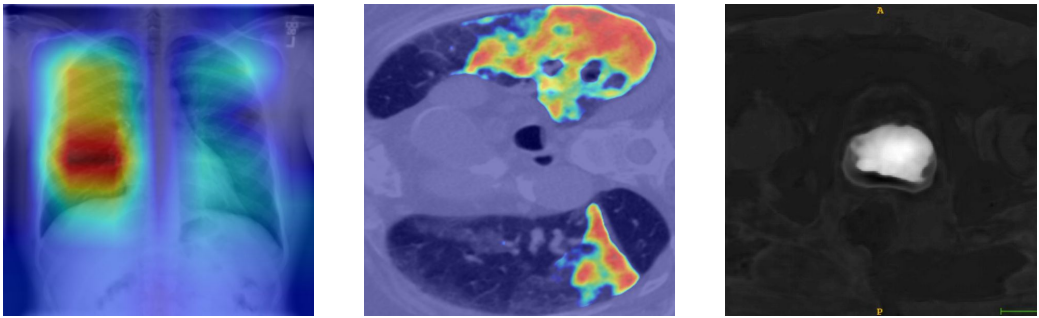

  \centering
  \begin{minipage}[b]{\x\textwidth}
    \includegraphics[width=\textwidth]{images/class_2D_gcam.jpg}
  \end{minipage}
  \hspace{0.5cm}
  \begin{minipage}[b]{\x\textwidth}
    \includegraphics[width=\textwidth]{images/seg_2D_gcam.jpg}
  \end{minipage}
    \hspace{0.5cm}
  \begin{minipage}[b]{\x\textwidth}
    \includegraphics[width=\textwidth]{images/seg_3D_gcam.jpg}
  \end{minipage}
    \caption{The resulting Grad-CAM attention maps from the input images.}
  \label{fig:gcam}
\end{figure}

Grad-CAM \cite{gcam} works by first propagating the input through the entire model. In a second step a desired class in the output is isolated by setting every other class to zero. The output of this isolated class is then backpropagated through the model up to the desired layer. Here the layer gradients are extracted and together with the feature maps of the same layer the attention map is computed. The result is a heatmap-like image of the attention at the desired layer as shown in figure \ref{fig:gcam}. The approach of generating an attention map from a specific preferably high layer gives a good compromise between high-level semantics and detailed spatial information. Furthermore, by isolating a specific class Grad-CAM becomes class discriminant.

\subsection{Guided Backpropagation}
\begin{figure}[!ht]
  \centering
  \begin{minipage}[b]{\x\textwidth}
    \includegraphics[width=\textwidth]{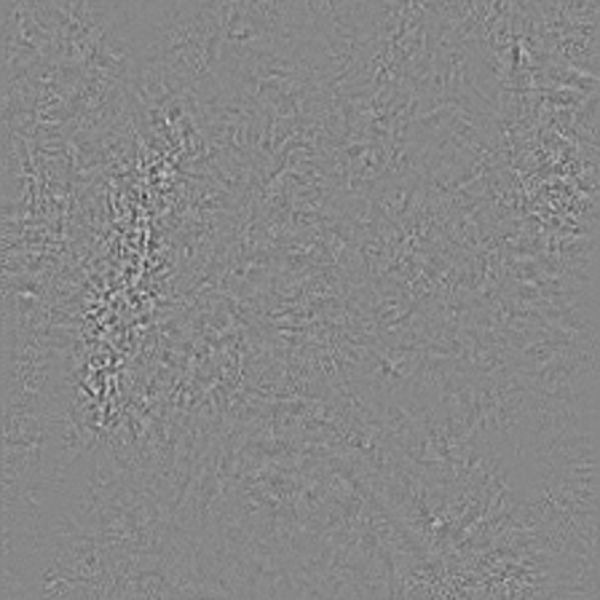}
  \end{minipage}
  \hspace{0.5cm}
  \begin{minipage}[b]{\x\textwidth}
    \includegraphics[width=\textwidth]{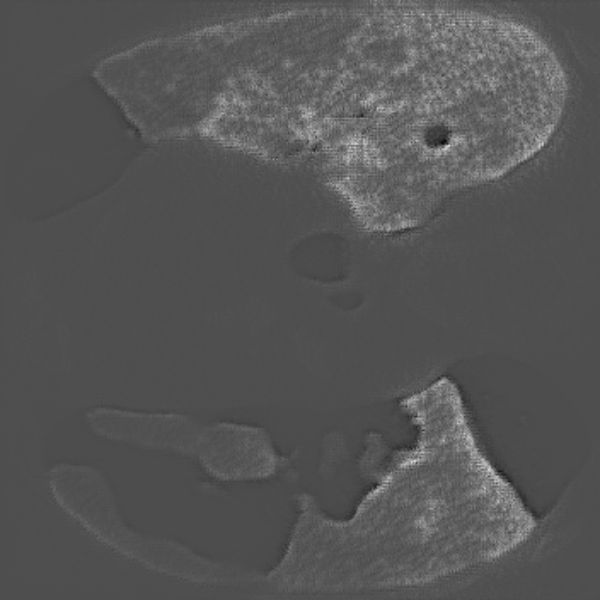}
  \end{minipage}
    \hspace{0.5cm}
  \begin{minipage}[b]{\x\textwidth}
    \includegraphics[width=\textwidth]{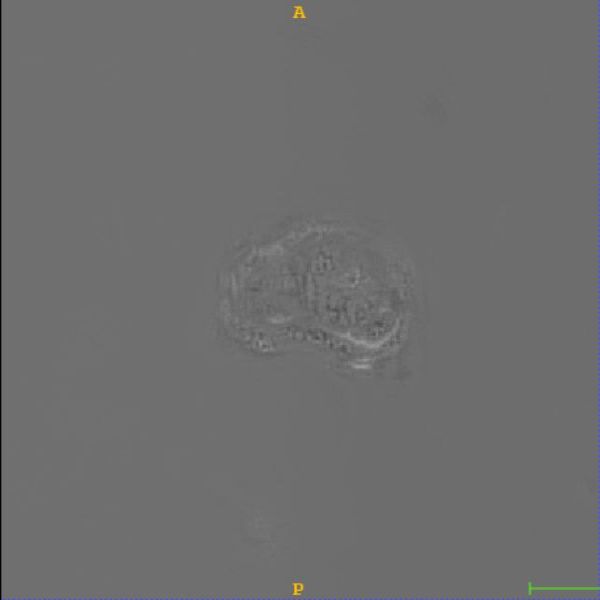}
  \end{minipage}
    \caption{The resulting Guided Backpropagation attention maps from the input images.}
  \label{fig:gbp}
\end{figure}

Guided Backpropagation was first introduced in \cite{gbp} and works by first propagating the input through the entire model similar to Grad-CAM. In a second step the output is then backpropagated through the entire model. However only the non-negative gradients are passed to the next layer as negative gradients correspond to suppressed pixels deemed not relevant by the authors. The result is a noise-like image depicting the model attention as shown in figure \ref{fig:gbp}. The advantage of Guided Backpropagation is that the attention is pixel-precise. The downsides are that it is neither class nor layer discriminant.

\subsection{Guided Grad-CAM}
\begin{figure}[!h]
  \centering
  \begin{minipage}[b]{\x\textwidth}
    \includegraphics[width=\textwidth]{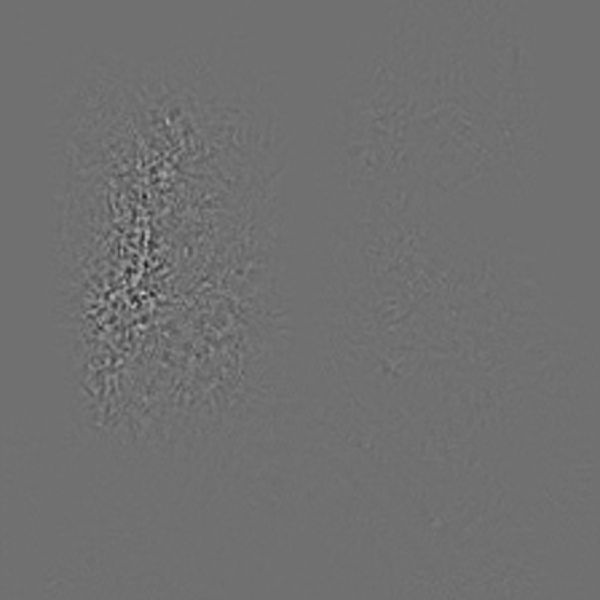}
  \end{minipage}
  \hspace{0.5cm}
  \begin{minipage}[b]{\x\textwidth}
    \includegraphics[width=\textwidth]{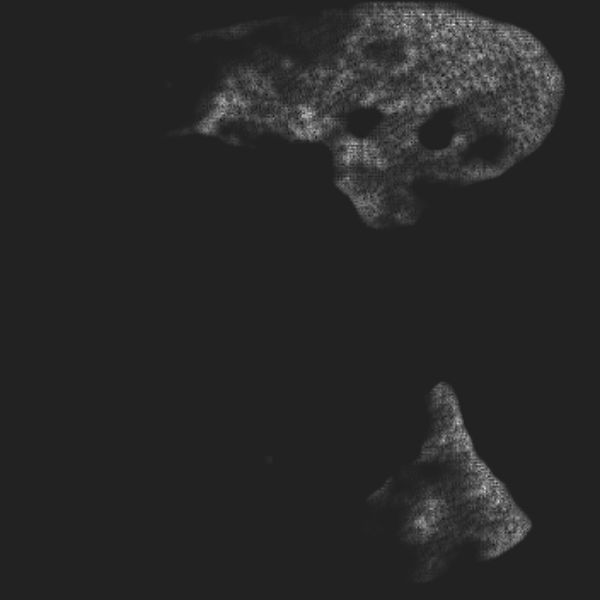}
  \end{minipage}
    \hspace{0.5cm}
  \begin{minipage}[b]{\x\textwidth}
    \includegraphics[width=\textwidth]{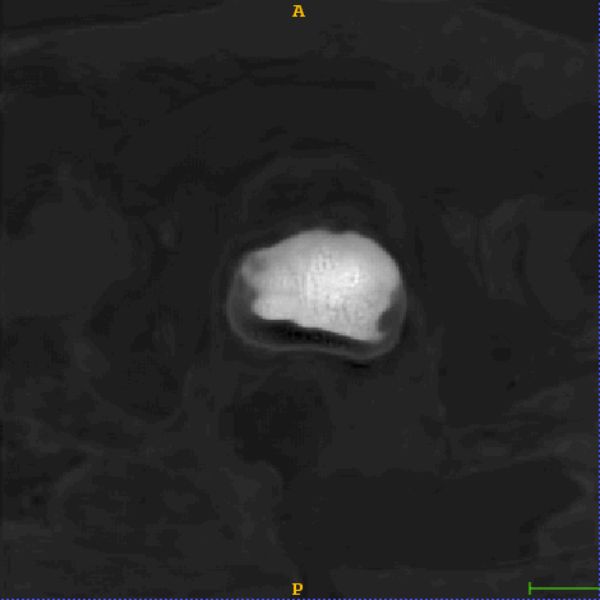}
  \end{minipage}
    \caption{The resulting Guided Grad-CAM attention maps from the input images.}
  \label{fig:ggcam}
\end{figure}

Another backend presented in \cite{gcam} is Guided Grad-CAM which is a combination of Guided Backpropagation and Grad-CAM in an effort to combine the best of both approaches. When generating attention maps with both backends the resulting attention maps can be combined through simply multiplying them element-wise. The result is a noise-like class and layer discriminant pixel-precise attention map as shown in figure \ref{fig:ggcam}. The only downside of Guided Grad-CAM is the need of performing backpropagation two times.

\subsection{Grad-CAM++}
\begin{figure}[!ht]
  \centering
  \begin{minipage}[b]{\x\textwidth}
    \includegraphics[width=\textwidth]{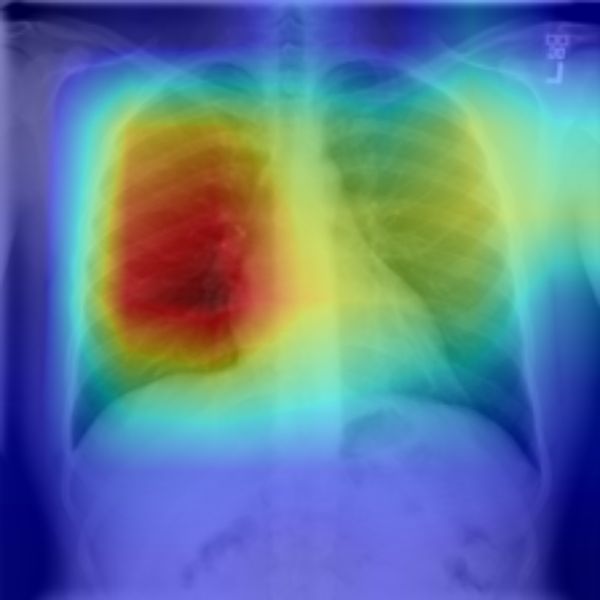}
  \end{minipage}
  \hspace{0.5cm}
  \begin{minipage}[b]{\x\textwidth}
    \includegraphics[width=\textwidth]{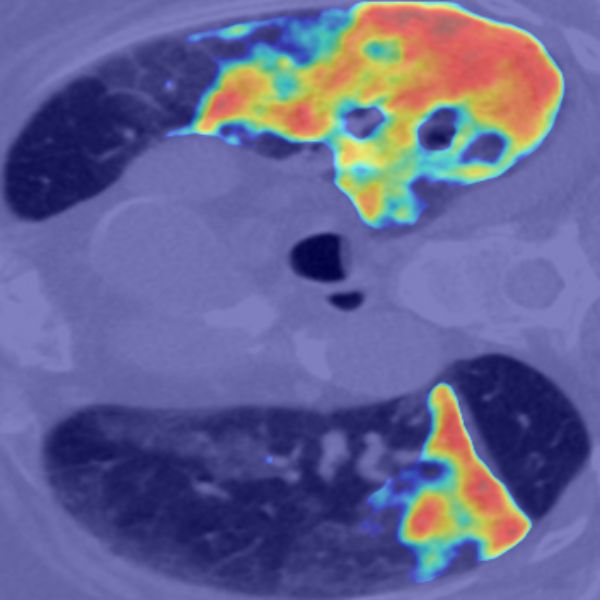}
  \end{minipage}
    \hspace{0.5cm}
  \begin{minipage}[b]{\x\textwidth}
    \includegraphics[width=\textwidth]{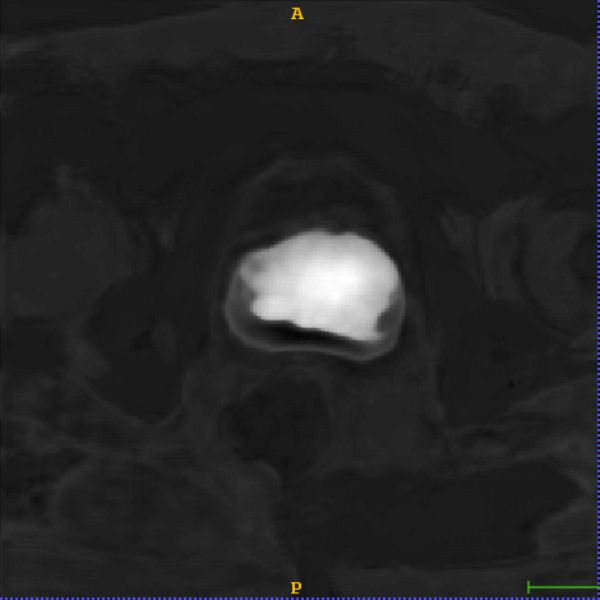}
  \end{minipage}
    \caption{The resulting Grad-CAM++ attention maps from the input images.}
  \label{fig:gcampp}
\end{figure}

Grad-CAM++ is an extension of Grad-CAM introduced in \cite{gcampp}. It differs to vanilla Grad-CAM in that it weights the gradients before combining them with the feature maps resulting in more precise attention maps, especially when dealing with multiple instances of the same class in an image according to the authors. Examples of these attention maps are shown in figure \ref{fig:gcampp}.

\printbibliography

@inproceedings{pytorch,
  title={PyTorch: An imperative style, high-performance deep learning library},
  author={Paszke, Adam and Gross, Sam and Massa, Francisco and Lerer, Adam and Bradbury, James and Chanan, Gregory and Killeen, Trevor and Lin, Zeming and Gimelshein, Natalia and Antiga, Luca and others},
  booktitle={Advances in Neural Information Processing Systems},
  pages={8024--8035},
  year={2019}
}

@article{gbp,
  title={Striving for simplicity: The all convolutional net},
  author={Springenberg, Jost Tobias and Dosovitskiy, Alexey and Brox, Thomas and Riedmiller, Martin},
  journal={arXiv preprint arXiv:1412.6806},
  year={2014}
}

@inproceedings{gcam,
  title={Grad-cam: Visual explanations from deep networks via gradient-based localization},
  author={Selvaraju, Ramprasaath R and Cogswell, Michael and Das, Abhishek and Vedantam, Ramakrishna and Parikh, Devi and Batra, Dhruv},
  booktitle={Proceedings of the IEEE international conference on computer vision},
  pages={618--626},
  year={2017}
}

@inproceedings{gcampp,
  title={Grad-cam++: Generalized gradient-based visual explanations for deep convolutional networks},
  author={Chattopadhay, Aditya and Sarkar, Anirban and Howlader, Prantik and Balasubramanian, Vineeth N},
  booktitle={2018 IEEE Winter Conference on Applications of Computer Vision (WACV)},
  pages={839--847},
  year={2018},
  organization={IEEE}
}

@article{nnunet,
  title={nnu-net: Self-adapting framework for u-net-based medical image segmentation},
  author={Isensee, Fabian and Petersen, Jens and Klein, Andre and Zimmerer, David and Jaeger, Paul F and Kohl, Simon and Wasserthal, Jakob and Koehler, Gregor and Norajitra, Tobias and Wirkert, Sebastian and others},
  journal={arXiv preprint arXiv:1809.10486},
  year={2018}
}

@misc{covidnet,
    title={COVID-Net: A Tailored Deep Convolutional Neural Network Design for Detection of COVID-19 Cases from Chest Radiography Images},
    author={Linda Wang, Zhong Qiu Lin and Alexander Wong},
    year={2020},
    eprint={2003.09871},
    archivePrefix={arXiv},
    primaryClass={cs.CV}
}

@article{infnet,
title={Inf-Net: Automatic COVID-19 Lung Infection Segmentation from CT Images},
author={Fan, Deng-Ping and Zhou, Tao and Ji, Ge-Peng and Zhou, Yi and Chen, Geng and Fu, Huazhu and Shen, Jianbing and Shao, Ling},
journal={IEEE TMI},
year={2020}
}

@article{covid19,
  title={COVID-19 image data collection},
  author={Joseph Paul Cohen and Paul Morrison and Lan Dao},
  journal={arXiv 2003.11597},
  url={https://github.com/ieee8023/covid-chestxray-dataset},
  year={2020}
}

@article{prostate,
  title={A large annotated medical image dataset for the development and evaluation of segmentation algorithms},
  author={Simpson, Amber L and Antonelli, Michela and Bakas, Spyridon and Bilello, Michel and Farahani, Keyvan and Van Ginneken, Bram and Kopp-Schneider, Annette and Landman, Bennett A and Litjens, Geert and Menze, Bjoern and others},
  journal={arXiv preprint arXiv:1902.09063},
  year={2019}
}

\end{document}